\documentclass[preprint,12pt,authoryear]{elsarticle}

\usepackage{times}
\usepackage{soul}
\usepackage{url}
\usepackage[hidelinks]{hyperref}
\usepackage[utf8]{inputenc}
\usepackage[small]{caption}
\usepackage{amsmath}
\usepackage{amsthm}
\usepackage{algorithm}
\usepackage{algorithmic}
\usepackage[switch]{lineno}
\usepackage{subcaption}
\usepackage{graphicx}
\usepackage{amsmath}
\usepackage{amssymb}
\usepackage{xcolor}
\usepackage{bm}
\usepackage{booktabs}
\usepackage{multirow} 
\usepackage{tabularx}
\usepackage{bbm}

\journal{Engineering Applications of Artificial Intelligence}

\begin{document}

\begin{frontmatter}

\title{WEITS: A Wavelet-enhanced residual framework for
interpretable time series forecasting}

\author[1]{Ziyou Guo}
\ead{guozy22@mails.jlu.edu.cn}
\author[2]{Yan Sun}
\ead{yansun414@gmail.com}
\author[1]{Tieru Wu\corref{cor}}
\ead{wutr@jlu.edu.cn}
\address[1]{School of Artificial Intelligence, Jilin University, Changchun, China}
\address[2]{H. Milton Stewart School of Industrial and Systems
Engineering, Georgia Institute of Technology, Atlanta, United States}
\cortext[cor]{Corresponding author}
\begin{abstract}
%% Text of abstract
  Time series (TS) forecasting has been an unprecedentedly popular problem in recent years,
with ubiquitous applications in both scientific and business fields. Various approaches have been
introduced to time series analysis, including both statistical approaches and deep neural networks.
Although neural network approaches have illustrated stronger ability of representation than statistical methods, they struggle to provide sufficient interpretablility, and can be too complicated
to optimize. In this paper, we present WEITS, a frequency-aware deep learning framework that
is highly interpretable and computationally efficient. Through multi-level wavelet decomposition,
WEITS novelly infuses frequency analysis into a highly deep learning framework. Combined with a forward-backward residual architecture, it enjoys both high representation capability and statistical interpretability. Extensive experiments on real-world datasets have demonstrated competitive
performance of our model, along with its additional advantage of high computation efficiency.
Furthermore, WEITS provides a general framework that can always seamlessly integrate with
state-of-the-art approaches for time series forecast.
\end{abstract}


\end{frontmatter}

%% main text
\section{Introduction}

Time series forecasting has been an important problem, required by many real-world applications, such as stock prediction \citep{idrees2019prediction}, inventory control \citep{aviv2003time} and disease modeling \citep{allard1998use}. It is of considerable scientific and financial impact to provide accuracy forecast based on historical series. Classical statistical tools, often employed with ensembling techniques, are still the workhorse in many forecast scenarios due to its interpretability and simplicity. With the explosive popularity of Transformers \citep{vaswani2017attention} in computer vision and natural language processing \citep{wolf2020transformers} since 2017, neural forecasting methods for time series has also evolved drastically, especially for long-horizon forecasts. 

However, despite the roaring success in the realms of long series forecast, deep learning techniques still struggle to completely outperform classical time series forecast approaches in multiple scenarios. Besides, lack of interpretability is still constraining deep learning techniques in certain fields, such as healthcare and business scenarios, despite of some literature in sought of interpretability of deep neural networks \citep{chefer2021transformer}. N-BEATS model \citep{nbeats} proposes a deep neural structure based on backward and forward residual links and a deep stack of fully connected layers, and have achieved competitive performance in multiple tasks. However, there are several disadvantages in the structural design: under its generic configuration, the model often fails to provide interpretable outputs from each stack, while interpretable configuration limits the number of stacks and requires careful selection of function basis. Besides, the structure of blocks are identical, which fails to capture multi-rate signals well. Therefore, a general and interpretable deep framework without handcrafted tuning is still strongly desirable. 

In this paper, we present a \textbf{W}avelet \textbf{E}nhanced deep framework for \textbf{I}ntepretable \textbf{T}ime \textbf{S}eries forecast (WEITS), for time series forecast with long-term dependency. By integrating original function spaces and wavelet induced spaces, WEITS combines two schools of thought to enhance interpretability while providing an expressive deep network. We also provide a parallelizable variation of WEITS that is computationally effective while still yielding accurate forecast. Furthermore, our method explores beyond a specific model structure, and can be viewed as an investigation of general frameworks with combination of arbitrary components, such as Transformer-based models.

\section{Summary of contributions}
As the first paper that combines frequency analysis (MDWD) and double residual framework with large deep learning models, our paper highlights several key contributions.  
\paragraph{Wavelet decomposition for interpretability}
Wavelet decomposition functions as a highly efficient tool for feature engineering, and can decompose a time series into a group of sub-series ranking by frequency. Through integrating wavelet transformation into a double residual framework, our method resembles existing approaches by decomposing series into long-term trend and high variability model, while the introduction of wavelet fully decouples the series without additional computation cost. Further discussion on the interpretability of the double residual network can be found in \cite{nbeats}. 
\paragraph{Residual stacking with multi-resolution}
Through the introduction of hierarchical wavelet decomposition and convolution mechanism, WEITS acquires the ability to capture information from an extremely long sequence history. Combined with the residual framework, WEITS can simultaneously output interpretable multi-resolution forecast from each stack. 
\paragraph{Generalizability of framework}
We propose a modified hierarchical double residual network, which explores beyond a specific model structure, and can be viewed as a general framework with combination of a wide range of forecast methods.

\section{Related works}
\paragraph{Multilevel wavelet decomposition} 
Wavelet decomposition is well known as a time-frequency analysis tool in time series \citep{percival2000wavelet}, and is usually applied for data pre-processing before deep networks \citep{wang2018multilevel}. Various types of wavelet have been applied in time series literature, including Haar wavelet ('db1') \citep{stankovic2003haar}, Daubachies wavelet \citep{daubechies1992ten} and sym4 wavelets \citep{sridhar2014wavelet}. Multilevel discrete decomposition (MDWD) can further extract time and frequency features of a time series in multiple levels ranging from high frequencies to low frequencies. Moreover, there are other works that achieve similar multi-rate sampling purpose \citep{ghysels2007midas}, and has been applied in deep models to enhance interpretability \citep{wang2018multilevel, singhal2022fusion, lai2018modeling}. 
\paragraph{Neural forecasting}
In the last decade, with the prosperity of deep learning concepts, various deep structures have been applied in time series analysis, and have achieved promising performance in versatile scenarios, with the examples of e-commerce retail \citep{qi2019deep}, financial forecast \citep{sezer2020financial} and disease modeling \citep{sun2021manifold}. The highlighted performance of deep learning methods in numerous forecast contests \citep{makridakis2020m4} have also drawn the attention of academic community. Earlier works adopt deep neural networks \citep{assaf2019explainable}, RNN \citep{yu2017long, wen2017multi, rangapuram2018deep, maddix2018deep}, LSTM \citep{cao2019financial}, TCN \citep{bai2018empirical}) with numerous variants to time series analysis and achieved evident success. 

The introduction of Transformer \citep{vaswani2017attention} has enlightened many following studies in time series analysis especially in long-horizon forecast \citep{das2023long}, which have dominated the landscape in the recent years. Most of the Transformer-based approaches (Autoformer \citep{wu2021autoformer}, LogTrans \citep{li2019enhancing}, Reformer \citep{kitaev2020reformer}, Informer \citep{zhou2021informer}, FedFormer \cite{zhou2022fedformer}) focus on improving computation efficiency and memory usage to improve prediction speed in long-horizon forecast. Although Transformer-based approaches achieve dominant success in long-sequence forecast, they still fail to dominate in the classical time series forecast tasks. Meanwhile, newest literature \citep{nie2022time, zeng2023transformers} has brought the effect of Transformer architecture in time series forecast to controversy. 

\paragraph{Interpretable forecasting}
Interpretable forecast \citep{lu2023importance} is an important concept within clinical ML as the provision of an interpretable explanation can aid in decision-making using ML models when the model performance is not perfect (\cite{hine2023blueprint}). Statistical methods are still the workhorse of interpretable time series forecasting (\cite{makridakis2018m4}), due to rigorous mathematical derivation. Classical statistical methods including ARIMA model, ARCH and GARCH model and Holt-Winters statistical seek to identify time series as auto-regressive variables, with trend and seasonality components. Spectral analysis decomposes time series into signals of different frequency, including Fourier transform, wavelet filtering \citep{joo2015time} and Koopman operator \citep{surana2020koopman}. 
Classical approaches have shown strong predictive power, but are gradually struggling to deal with complicated scenarios nowadays. Therefore, exploring the interpretability of deep learning methods have gained popularity in recent years \citep{ismail2020benchmarking}. Some literature focus on developing interpretability for model structure, such as  LSTM \citep{guo2019exploring, guo2018interpretable} and attention mechanism \citep{chefer2021transformer}. Interpretability can also be enhanced through combination of deep models with classical methods. For example, DeepAR \citep{salinas2020deepar} proposed recurrent neural network with auto-regressive properties. N-Beats \citep{nbeats} and N-Hits model \citep{challu2022n} utilized residual network to decompose time series into trend and seasonability. \citep{wang2018multilevel} proposed a wavelet-based LSTM network to learn low and high frequency characteristics of time series. 

\section{Methodology}
\begin{figure}[h!]
\vspace{.3in}
     \centering
         \includegraphics[width=4.4in]{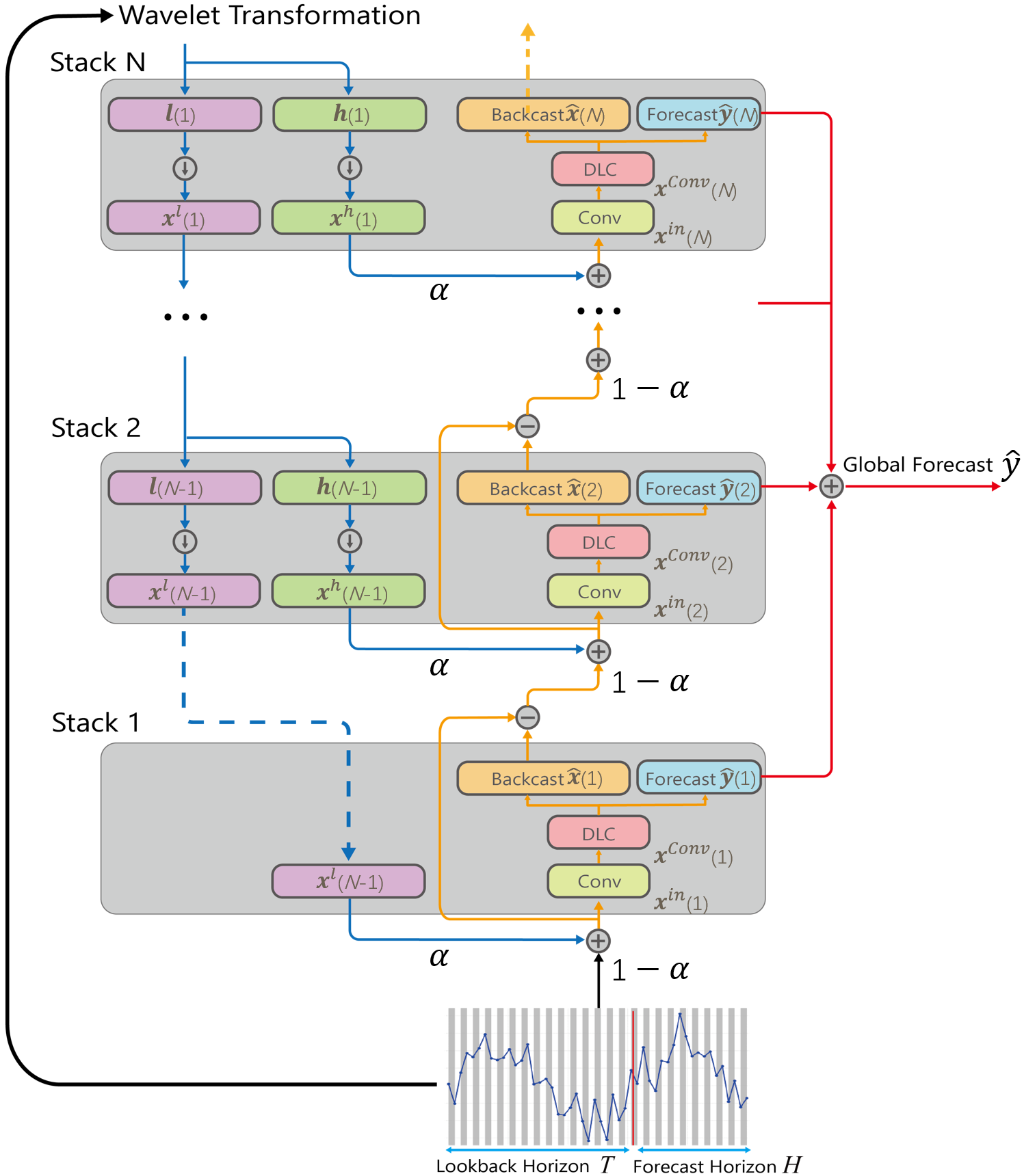}
\vspace{.3in}
\caption{Main structure of WEITS. The input to each stack is the residual from the last stack infused with a corresponding sub-series from wavelet decomposition (MDWD). Then the input goes through a dilated convolutional layer to the stack component (DLC) which generates partial backcast and forecast of the series. The global forecast is generated by summing up stack forecasts.}
\label{fig:main-structure}
\end{figure}

Our architecture design relies on several key principles. First, we provide a general framework that incorporates  time series forecast methods as compartmental operators, while still maintaining simple and explainable structure. Second, the architecture shall not include any series-specific pre-processing, but shall automatically infuse interpretability into model inputs. Besides, the model should be well-established in a human interpretable multi-scale architecture. The main WEITS structure is depicted in Figure \ref{fig:main-structure}. In the following sections we discuss how the components in the architecture can achieve all advantages above in detail.

\subsection{Overview: residual stacking with signal enhancement}
Residual networks \citep{he2016deep} enjoy a theoretical advantage of interpretability from an ensemble perspective \citep{veit2016residual}, and are widely applied in deep learning for increased performance and trainability. In the classical residual network structure, the input of the layers are either added to the layer output as the input of the next layer \citep{he2016deep}, or added to all following layers \citep{huang2017densely}. Double residual stacking \citep{challu2022n} further improves the interpretability of residual architectures by proposing two branches, one iterating over backcast of each layer and the other over all forecast predictions. However, as there is no guarantee that the residual network provides ideal decomposition unless the basis functions are handcrafted. Therefore, such architecture still often fails to provide the partial (stack) predictions that are easy to interpret. The infusion of wavelets effectively fixes the issue. Besides, the expression capacity of the model is further enhanced by harnessing the proportion of wavelet infusion. We also introduced a convolutional mechanism that is effective for capturing long-term dependency.

\subsection{Signal enhancement: MDWD}

We denote the uni-variate input observations as $\bm{x}=[a_1, ...,a_T] \in \mathbb{R}^{T}$. The task is to predict the value of series $\bm{y}=[a_{T+1}, ...,a_{T+H}] \in \mathbb{R}^{H}$ for a fixed horizon length $H$, given historical observations $\bm{x}$. For a forecast task with long-range dependency, $T \gg H$. For notation convenience, letters with bold face will be used to denote vectors, and letters with default font will refer to real numbers.

We first dive into the details of wavelet decomposition. The basic structure of Multilevel Discrete Wavelet Decomposition (MDWD) is shown in the left panel in Fig.\ref{fig:main-structure}. Denote the low and high sub-series generated in the $n$-th level as $\bm{x}^{l}(n)$ and $\bm{x}^{h}(n)$, for $n=1,2...,N-1$ where $N$ is the number of stacks which will be introduced later. For wavelet transformation, WEITS uses a low pass filter $\bm{l}=[l_1,...,l_K]$ and a high pass filter $\bm{h}=[h_1,...,h_K]$, where $K \ll T$. We convolute low frequency sub-series of the upper level:

\begin{align}
\begin{split}
\begin{gathered}
    l_t(i+1)=\sum^K_{k=1} a_{t-k+1}^{l}(i)\cdot l_k \\
    h_t(i+1)=\sum^K_{k=1} a_{t-k+1}^{l}(i)\cdot h_k 
\end{gathered}
\end{split}
\end{align}

\begin{figure}[h!]
     \centering
         \includegraphics[width=4in]{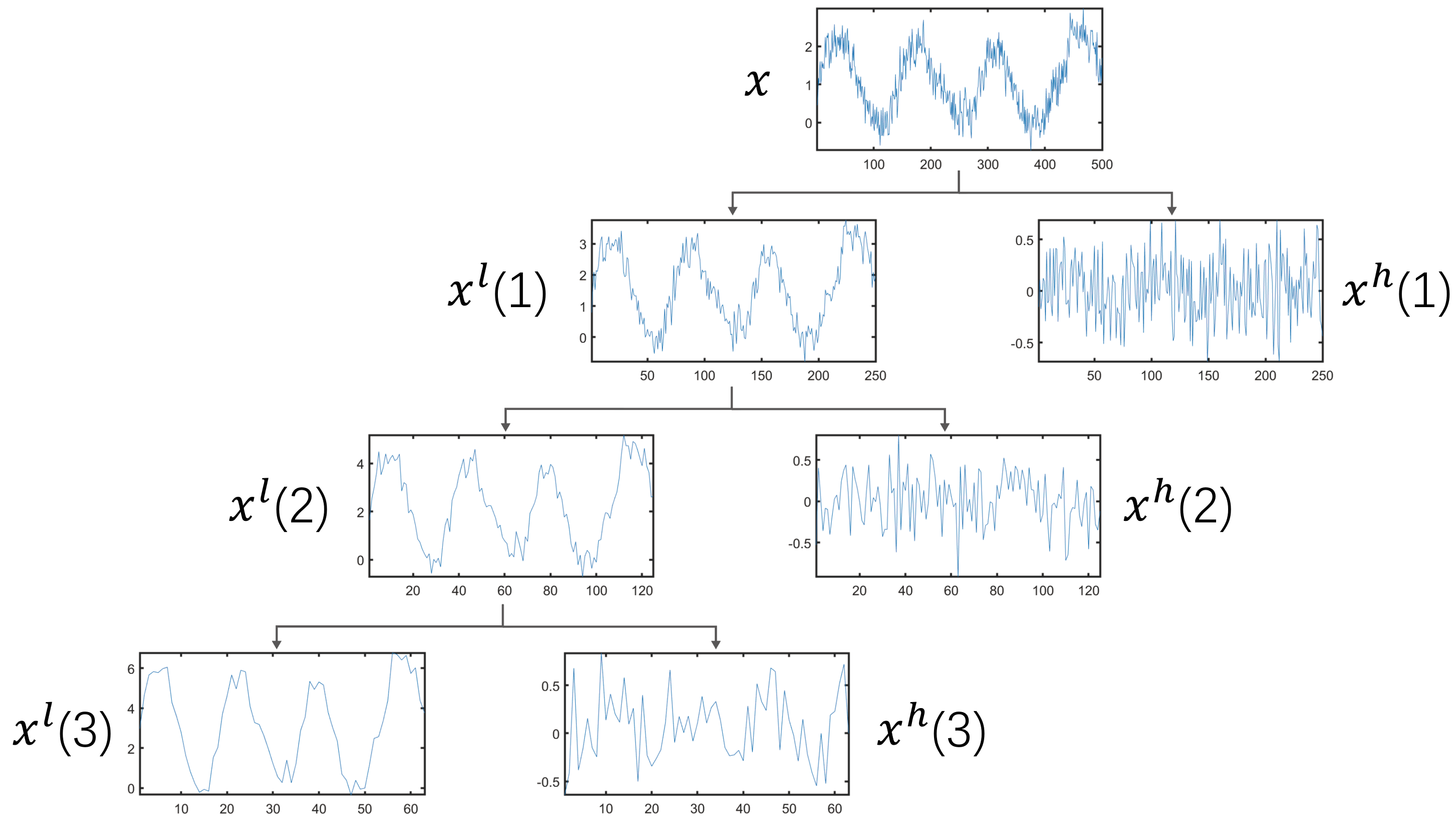}
    \vspace{.3in}
\caption{Illustration of MDWD. In each level the series is decoupled into a high- and low-frequency sub-series.}
\label{fig:mdwd}
\end{figure}

Where $a_{t}^{l}(i)$ is the $t$-th element of low frequency sub-series in the $i$-th level series $\bm{x}(i)$, and $\bm{x}^l(0)$ is the original series. An illustration of MDWD is presented in Fig.\ref{fig:mdwd}. Based on the notations above, the original series will be decomposed into sub-series with different frequancy ranges, and sub-series in the $i$-th level $\bm{x}^{l}(i)$ should capture long-term trends when $i$ is large. Following multilevel discrete wavelet decomposition technique \citep{mallat1989theory}, low-frequency sub-series $\bm{x}^l(i)$ and high-frequency sub-series $\bm{x}^h(i)$ are reconstructed from the 1/2 down-sampling of $\bm{l}(i)$ and $\bm{h}(i)$ respectively, and have the same dimension as $\bm{x}$. Note that infusing wavelet sub-series does not introduce series-specific processing, because once the type of wavelet function is settled, the decomposition of the series is very fast and will be determined throughout training. The implementation of MDWD can be found in Python PyWavelets \citep{lee2019pywavelets} and Matlab Wavelet Toolbox \citep{misiti1996wavelet}.

The structure of building components of each stack is described in the right panel of Fig.\ref{fig:main-structure}. In the first stack, instead of the original input, we enhance the input data by infusing patterns of certain frequencies that we would like the stack to learn by using a convex combination of wavelet signals and residuals from the previous stack:

\begin{align}
\begin{split}
\small
     \bm{x}^{in}(i)=
     &\begin{cases}
\alpha  \bm{x}^l(N-1) + (1-\alpha) \bm{x}, \\ (i=1) \\
\\
\alpha  \bm{x}^h(N-i+1) + (1-\alpha) (\bm{x}^{in}(i-1) - \bm{\hat x}(i))),  
\end{cases}
\\
&\ \ \ (i=2,3,...,N)
\end{split}
\end{align}

Where $\bm{x}^{in}(i)$ is the input of the $i$-th stack. $\alpha$ is a tunable parameter that harnesses how much signal the deep model shall learn from the wavelet patterns. To avoid induced bias of inference, $\alpha$ shall be shared among all stacks. Then the inputs pass through stack $i$ which generates a forward forecast $\bm{\hat y}_i=[\hat a_{T+1},...,\hat a_{T+H}]$, along with the stack's best estimate of $\bm{x}^{in}(i)$, denoted as 'backcast' $\bm{\hat x}(i)$. The derivation of backcast and forecast will be discussed in Section \ref{sec:blocks}.

\subsection{Multi-resolution sampling}

\begin{figure*}[h!]
     \centering
         \includegraphics[width=5.2in]{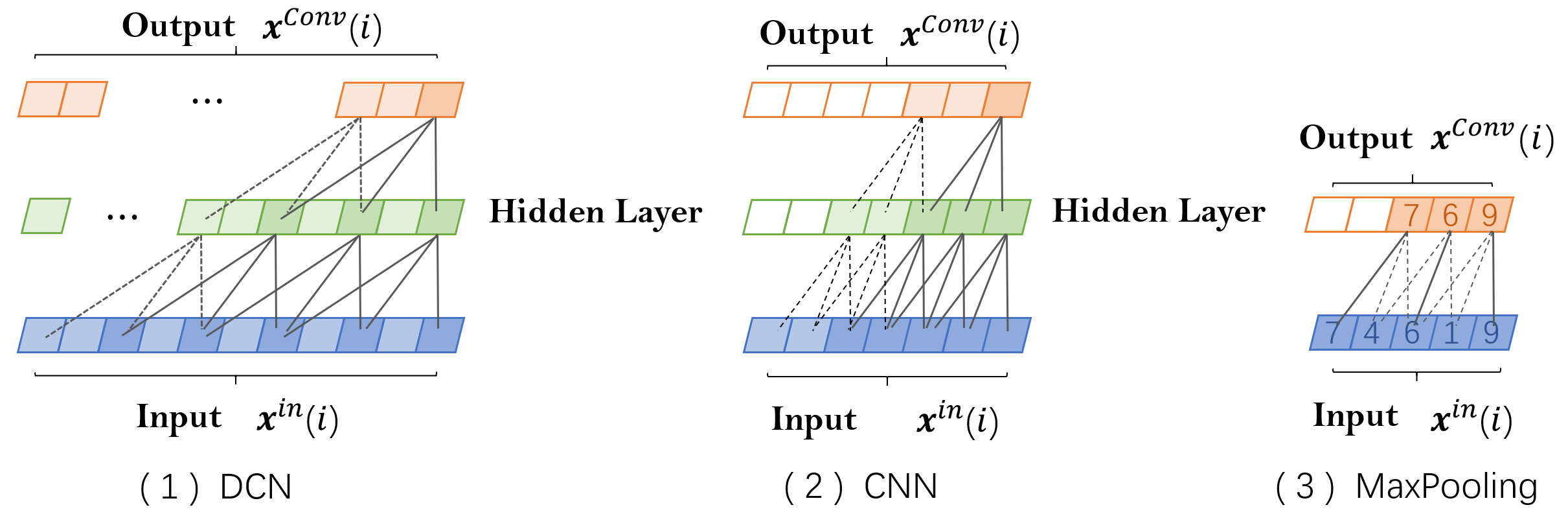}
    \vspace{.3in}
\caption{WEITS can utilize multiple structures as convolutional components within each stack, including DCN, CNN and MaxPooling.}
\label{fig3}
\end{figure*}

To focus on analyzing the components of the stack input with different resolution scales, we introduce a dilated 1D-convolution layer with kernel size $k_i$ directly after stack $i$ receives the input $\bm{x}^{in}_i$, described in Figure \ref{fig3}.  Larger kernel size $k_i$ will filter out high frequency components (local patterns) and force the stack to focus on low frequency (global) patterns. Inclusion of a convolution layer in each stack also helps to reduce the size of inputs to each stack and limit the number of learnable parameters in the network. Moreover, the multi-resolution mechanism can be seen as a regularization approach, which alleviates the effect of over-fitting, while still keeping the original receptive field. In the WEITS model, stacks with larger index $i$ is designed to capture higher frequency patterns. Therefore, the sequence of kernel size $k_1,...,k_N$ should be monotonically decreasing. Note that different from existing methods \citep{nhits}, the convolution layer is applied within each stack instead of each block. Such modification allows the components within the stack to be more flexible, and has been proved to be effective in the experiments (see Section \ref{sec:experiment}). We denote the outputs after convolution layer as $\bm{x}^{Conv}(i)$ for each stack $i$, with the same dimension of $\bm{x}$. Besides dilated convolutional neural network, we also explore the performance of different convolutional kernels, including MaxPool, AveragePool, and multi-layer CNN, with results shown in Section \ref{sec:different-cnn}. 

\subsection{Building blocks of stacks} \label{sec:blocks}
WEIT is a high-level network architecture without contraining the architectures of its inner stacks. To demonstrate the generalizability and good performance of WEITS, we choose two deep learning components (DLCs) for stack construction in this study: a multi-layer FC network with basis expansion, and a transformer-based component. However, the potential of WEITS goes beyond these architectures. In fact any algorithm that can generate both backcast and forecast (e.g., temporal convolutional neural networks (TCN)) can be utilized as system component in the stack. 
\subsubsection{Neural basis expansion}
The FCN components with neural basis expansion follow generic configuration from \cite{nbeats}, Within each stack there are $K$ blocks with identical structure. Denote the input and output to the $k$-th block in the $i$-th stack as $\bm{x}^{in}(i,k)$ and $\bm{x}^{out}(i,k)$ respectively. The operation within block can be described by the following equations:
\begin{align}
\begin{split}
\begin{gathered}
    \bm{h}(i,k) = \textbf{\textsc{NN}}_{i,k}(\bm{x}^{in}(i,k-1)-\bm{x}^{out}(i,k-1)) \\
    \bm{\theta}^{b}(i,k) = \textbf{\textsc{NN}}^b(\bm{h}(i,k)) \\
    \bm{\theta}^{f}(i,k) = \textbf{\textsc{NN}}^f(\bm{h}(i,k)))
\end{gathered}
\end{split}
\end{align}
Here $\bm{h}(i,k)$ is the intermediate output, and $\bm{\theta}^{f}(i,k)$ and $\bm{\theta}^{b}(i,k)$ are the forward and backward predictors of expansion coefficients. $\textsc{NN}^b(\bm{h}(i,k))$ and $\textsc{NN}^f(\bm{h}(i,k)))$ are two separate MLPs that generate predictions of expansion coefficients $\bm{\theta}^f(i,k)$ and $\bm\theta^b(i,k)$ respectively. By definition,
\begin{align}
\begin{split}
    \bm{x}^{in}(i,0)=\bm{x}^{Conv}(i), \quad
    \bm{x}^{out}(i,0)=0 \quad \\ \text{(stack input to block 1)} \\
    \bm{x}^{in}(i+1)=\bm{x}^{in}(i,K-1)-\bm{x}^{out}(i,K) \quad \\ \text{(output to the next stack)}
\end{split}
\end{align}

We set the last layer of each block as a linear projection of coefficients $\bm{\theta}^{b}(i,k)$ and $\bm{\theta}^{f}(i,k)$ to generate block outputs. The following operations to generate block backcast and forecast, as well as stack forecast, can be described as:
\begin{align}
\begin{split}
\begin{gathered}
    \bm{\hat x}(i,k)=\bm{W}^b(i,k) \bm{\theta}^b(i,k) + \bm{b}^b(i,k) \\
    \bm{\hat y}(i,k)=\bm{W}^f(i,k) \bm{\theta}^f(i,k) + \bm{b}^f(i,k) \\
    \bm{\hat x}(i)= \sum^K_{k=1} \bm{x}^{out}(i,k), \quad \bm{\hat y}(i)= \sum^K_{k=1} \bm{\hat y}(i,k), \\
    \quad \bm{\hat y}= \sum^N_{i=1} \bm{\hat y}(i) \quad \quad\quad\text{(global forecast)}
\end{gathered}
\end{split}
\end{align}
Linear projection layers are not only highly generalizable, but also requires no prior information about time series. Compared to existing approaches \citep{nbeats}, which rely on handcrafted selection of basis function vectors WEITS for each stack for interpretability, WEITS generates interpretable forecast immediately from each stack even under generic settings while still providing interpretable outputs, thanks to the infusion of sub-series to each stack. 

\subsubsection{Transformer-based components}
Along with existing approaches, Transformer-based architectures are also readily available to produce reliable backcast and forecast, with minimal modification of the model structure. However, optimization of multiple Transformer models with a fork structure can be extremely difficult. Therefore, more efficient variants of Transformers can be considered. For example, Informer \citep{zhou2021informer} has been a popular tool for long series forecast, which can be readily incorporated into our architecture. We modify the decoder in Informer to generate a concatenation of backcast and forecast predictions. Hyperparameters of the model are fine-tuned following \citep{chen2022towards}. 

\begin{figure}[h!]
     \centering
     \begin{subfigure}{0.5\textwidth} % Added width parameter here
         \vspace{.3in}
         \hspace{-1.5cm}\includegraphics[width=3.8in]{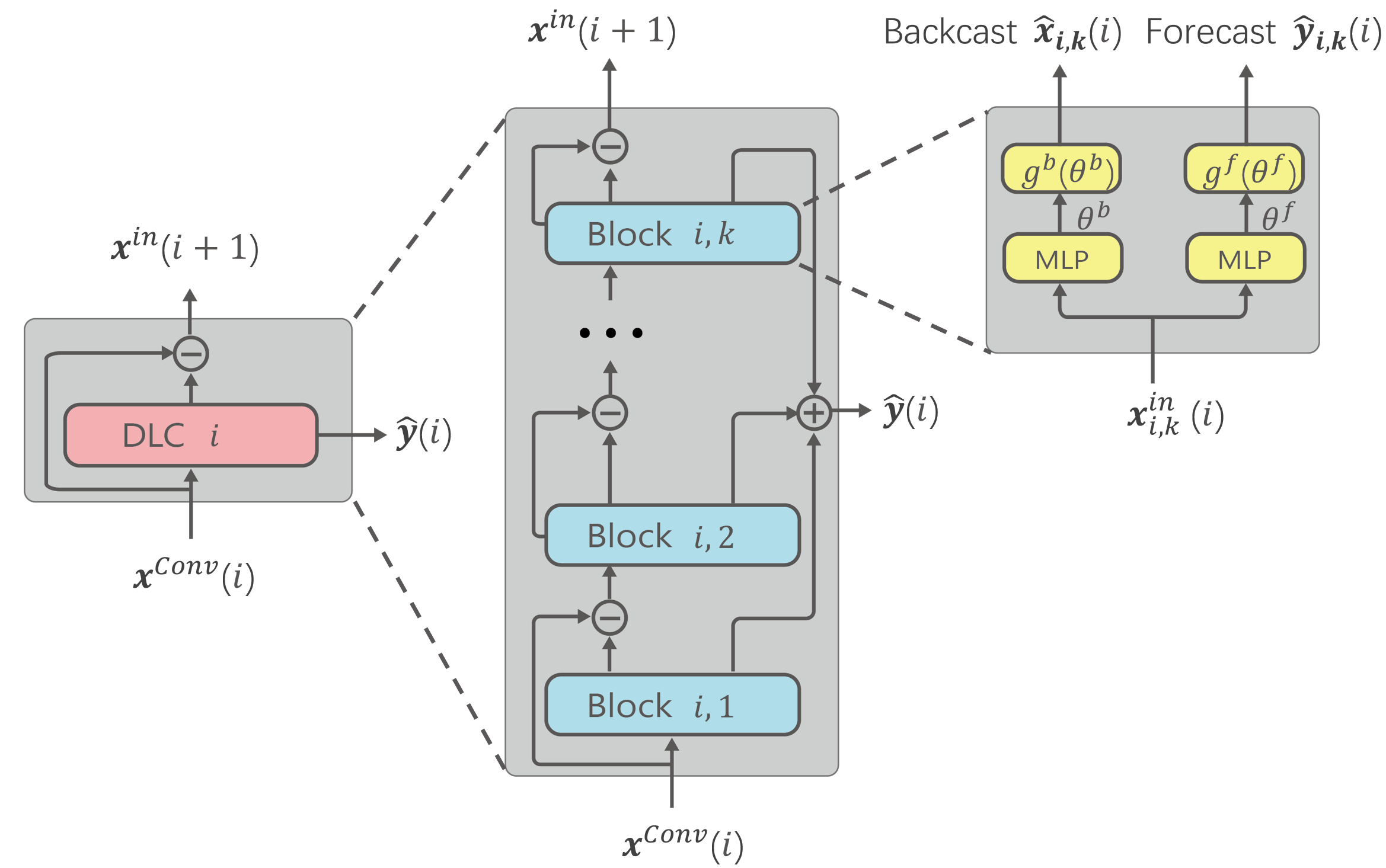}
         \caption{The structure of N-BEATS}
     \end{subfigure}
     
     \begin{subfigure}{0.5\textwidth} % Added width parameter here
         \vspace{.3in}
         \hspace{-1.5cm}\includegraphics[width=3.8in]{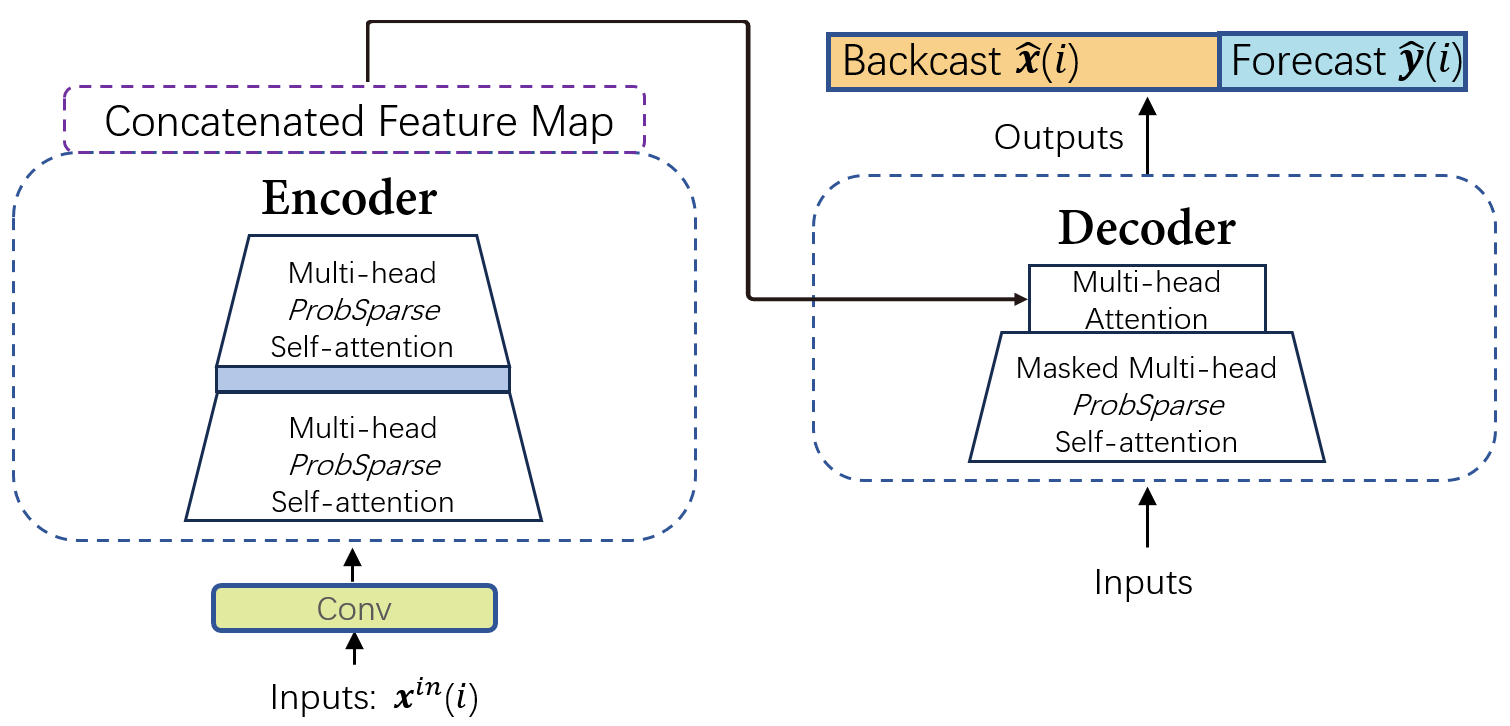}
         \caption{The structure of Informer}
     \end{subfigure}
     \vspace{.3in}
\caption{Structure illustration of FCN and Informer as stack component of WEITS.}
\label{fig5}
\end{figure}

\subsection{Ensembling}
Another significant advantage of our model is the high capacity of ensembling \citep{sagi2018ensemble}. Recent literature have demonstrated that ensembling is a much more powerful regularization method than other techniques (e.g. normalization or dropout) \citep{dietterich2000ensemble} in promoting prediction accuracy, because it incorporates additional sources of diversity to the original model. Our model relies on several ensembling techniques with different random initialization, utilizing 40 total models and performs a bagging procedure (\cite{galar2011review}). See Appendix Section 4 for details.

\section{Experimental results} \label{sec:experiment}

We demonstrate the performance of our model using four datasets: an SP-500 hourly dataset  and three other popular datasets for univariate forecast: ETTh1, ETTh2 \citep{zhou2021informer} and influenza-like illness dataset (ILI). We study the performance under multiple WEITS configurations: wavelet-infused FCN (WEITS-1), wavelet-infused Informer (WEITS-2), with key results presented in Table 1. Details of training and evaluation are discussed in Section \ref{sec:train}. We wrap up the experimental study through an ablation study in Section \ref{sec:ablation}. See \ref{dataset} for detailed introductions of these data sources.

\subsection{Training methodology} \label{sec:train}
Under FCN configuration, main model hyperparameters are set as layers per stack $K=5$, width and depth of the fully connected neural networks are determined through grid search for all neural networks $\textbf{\textsc{NN}}_{i,k}$, $\textbf{\textsc{NN}}^b$ and $\textbf{\textsc{NN}}^f$. Empirically a neural network with depth 3 and width 16 is recommended. Hyperparameters in the Informer component are configured with default settings. The default number of stacks $N$ in WEITS-1 is set as 4, which has been shown accurate while still being interpretable \citep{nbeats}, and $N=2$ for WEITS-2. A 3-layer TCN layer with h dilation factors $d = 1, 2, 4$ and filter size $k = 3$ is used in each stack when the observation length is less than 120, and a 4-layer TCN is used with for longer series input $d = 1, 2, 4, 8$. WEITS model is implemented in Pytorch \citep{paszke2019pytorch} and trained using Adam with a learning rate of 0.0001 over 100 epochs on each dataset with a early-stopping patience at 50 steps. The first 10\% of epochs are warm-up procedures, followed by a 0.001 linear decay of learning rate. The lookback window is fixed at 720 thanks to the natural adaptiveness of WEITS to long sequences (104 for the ILI dataset). We run our model on a single Nvidia RTX 4090 GPU, with a batch size of 128 for most datasets if possible, and  reduce the batch size to 16 if the sequence length is inadequate. The dropout rate is set at 0.1 to avoid over-fitting, while other settings remain as default, such as Xavier initialization and batch normalization \citep{ioffe2015batch}.

\subsection{Key results}

\subsubsection{Result overview}
We compare WEITS with following SoTA methodologies as benchmark methods: (1) N-BEATS \citep{nbeats}, (2) N-HiTS \citep{nhits}, (3) DeepAR \citep{salinas2020deepar} (4) auto-ARIMA \citep{hyndman2008automatic} (5) FEDformer \citep{zhou2022fedformer} (6) Informer \citep{zhou2021informer} and (7) Reformer \citep{kitaev2020reformer}. Method (1) to (4) are specially designed for univariate time series forecast \footnote{Official documentations of N-BEATS and N-HiTS achieve multivariate forecast through tensor flattening (concatenation).}, while method (5) to (7) are Transformer-based model that focus on multivariate, long-horizon forecast. Table 1 summarizes the experimental results on the datasets. WEITS has illustrated competitive performance across these datasets, achieving 7\% and 9\% decrease in MSE and MAE respectively. WEITS also enjoys a high computational efficiency under its parallel configuration, as well as the fast computation of wavelet decomposition. Additional results of WEITS under different configurations (parallel and heterogeneous) and more benchmark methods will be discussed in Appendix Section 1. Besides, extensive experiments are conducted to explore the forecast ability of WEITS in multivariate series and long-horizon tasks. See Appendix Section 2 for details. 

\begin{figure}[h!]
\vspace{.3in}
\centering
\includegraphics[width=5in]{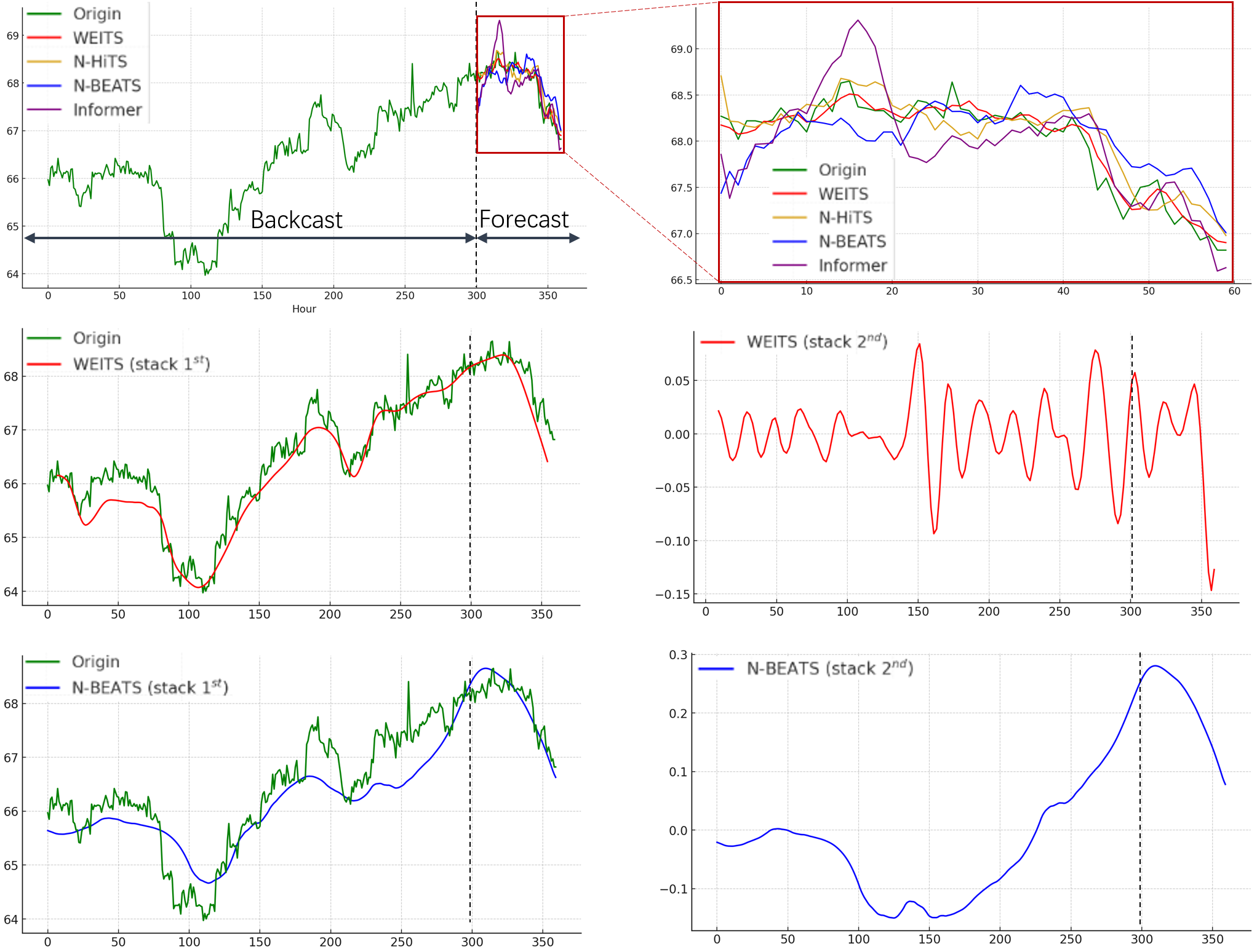}
\vspace{.3in}
\caption{Comparison of stack output from WEITS and N-Beats. WEITS provides an easily understandable stack output that shows both trends and volatility.}
\label{fig5}
\end{figure}

\begin{table*}
\footnotesize
\label{table1}
\caption{Comparison of model performance under different history / horizon length settings. Multivariate results with prediction horizon $L_P \in \{96, 192, 336, 720\}$ ($L_P \in {24, 36, 48, 60}$ for ILI). The best results are highlighted bold. Results for other models are sourced from the original literature or derived from additional experiments conducted for unreported. More benchmark comparison results can be found in Appendix Section 1.}

\begin{center}
\
\resizebox{\textwidth}{!}{\begin{tabular}{cc|ccccccccccccccc}
\toprule
 \multicolumn{2}{c}{Model}&\multicolumn{2}{c}{WEITS-1}&\multicolumn{2}{c}{WEITS-2}&\multicolumn{2}{c}{N-HiTs}&\multicolumn{2}{c}{N-Beats}&\multicolumn{2}{c}{DeepAR}&\multicolumn{2}{c}{Informer} \\
 Dataset & H &MSE&MAE&MSE&MAE&MSE&MAE&MSE&MAE&MSE&MAE&MSE&MAE \\
 \hline
\multirow{4}{*}{\rotatebox{90}{SP-500}}
&6  &\textbf{0.183}&\textbf{0.261}&0.194&0.271&0.221&0.270&0.294&0.327&0.308&0.367&0.296&0.311\\
&12 &\textbf{0.225}&\textbf{0.282}&0.254&0.301&0.241&0.297&0.305&0.322&0.336&0.379&0.329&0.362\\
&18 &\textbf{0.249}&\textbf{0.305}&0.257&0.309&0.283&0.314&0.349&0.386&0.370&0.416&0.383&0.477\\
&24 &\textbf{0.319}&\textbf{0.354}&0.324&0.368&0.351&0.402&0.448&0.509&0.409&0.436&0.402&0.436\\
\midrule
\multirow{3}{*}{\rotatebox{90}{ETTm2}}
&160
&0.129&0.252&0.137&0.257&\textbf{0.109}&\textbf{0.234}&0.146&0.239&0.172&0.288&0.287&0.346\\
&320
&\textbf{0.262}&0.318&0.295&0.354&0.278&\textbf{0.368}&0.289&0.341&0.316&0.437&0.385&0.482\\
&480
&0.451&0.498&0.436&0.508&\textbf{0.421}&\textbf{0.441}&0.484&0.523&0.902&0.842&0.761&1.267\\ 
\midrule
\multirow{3}{*}{ \rotatebox{90}{ECL}}
&160
&\textbf{0.169}&0.214&0.183&0.244&0.172&\textbf{0.203}&0.207&0.291&0.306&0.359&0.292&0.368\\
&320
&\textbf{0.187}&\textbf{0.245}&0.201&0.278&0.195&0.268&0.243&0.286&0.335&0.371&0.365&0.397\\
&480
&0.259&\textbf{0.330}&0.269&0.324&\textbf{0.244}&0.347&0.283&0.376&0.408&0.497&0.394&0.442 \\ 
\midrule
\multirow{3}{*}{\rotatebox{90}{Weather}}
&160
&\textbf{0.192}&0.276&0.225&0.251&0.290&\textbf{0.248}&0.298&0.329&0.313&0.350&0.327&0.391\\
&320
&\textbf{0.299}&\textbf{0.334}&0.317&0.346&0.307&0.352&0.388&0.406&0.403&0.389&0.449&0.478\\
&480
&0.372&0.394&0.390&0.421&\textbf{0.338}&\textbf{0.361}&0.426&0.407&0.601&0.628&0.783&0.845 \\
\bottomrule
\end{tabular}}
\end{center}
\end{table*}

\subsubsection{Interpretability of WEITS}
Forecasting practitioners often use the decomposition of time series into meaningful components, such as trend and seasonality \citep{cleveland1990stl, sax2018seasonal}, which is represented by a monotonic (or at least a slowly varying) function and a cyclical fluctuation, or into different volatility levels, which is important in stock prediction \citep{lv2022stock}. In different stacks of lower level of WEITS, the stack output is a low-frequency component, the so-called approximation, representing the trend, and higher level stacks yield a high-frequency component, the so-called detail, representing the detailed local features of the series. The decomposition of WEITS is similar to interpretable time series approaches which decomposes an original series into Intrinsic Mode Functions (IMFs) and a residual term \citep{torres2011complete}. This provides practitioners with information for further analysis. For example, in the Augmented Dickey Fuller (ADF) method \citep{cheung1995lag}, the low-volatility time series are classified as linear components, and high-volatility time series are classified as non-linear components, which enables the exploration of the these components with exogenous variables. Compared to existing methods \citep{nbeats}, the decomposition of WEITS is significantly more effective (see Table S7) and the decomposition of WEITS is automatic, eliminating the necessity of pre-defining the basis functions. Moreover, WEITS is more accurate in prediction with the introduction of double residual structure compared to vanilla wavelet decomposition only. 

\subsection{Ablation study} \label{sec:ablation}
We conduct ablation studies on the three primary modules within WEITS to showcase their specific
mechanisms for performance enhancement. We use the same setting for comparing different models
and fix $H=720$. Implementation details for using and transferring the modules are provided in
xxx. Additionally, we conducted further ablation studies on the details within the modules in xxx.
\subsubsection{Efficacy of wavelet infusion}
To explore whether infusing multilevel wavelet decomposition actually leads to performance improvement, we modify the experiment by fixing $\alpha$ at different levels. Table S2 shows the comparison of accuracy when different infusion levels. 

\subsubsection{Efficacy of double residual structure}
Wavelet decomposition has long been used in the literature, including \cite{wang2018multilevel, lai2018modeling}. We conduct an ablation experiment in Supplementary Table S7 to showcase the additional value compared to direct wavelet decomposition for time series forecasting. We see that the model accuracy significantly improves with the involvement of residual structure, confirming the validity of the structure design.

\subsubsection{Number of stacks}
In WEITS architecture, the number of stacks is determined by the number of wavelet decomposition levels. Table S4 in Appendix shows the relation between number of stacks and prediction error. Although increasing the number of stacks leads to higher performance, the interpretability of stack output might be weakened. For practical usage, 4 to 6 stacks are typically recommended, as higher level value may lead all coefficients to experience boundary effects.

\subsubsection{Design of block architecture}
We extensively investigate whether variations of block architectures may further promote forecast accuracy, or improve computation efficiency by exploring multiple designs of block structure. WEITS-3 substitudes the FCN structure in the first stack with an Informer component. WEITS-4 achieves a fully parallel capability by using parallel configuration and setting $\alpha=1$ , which can be seen as a super learner consisting sub-models focusing on different resolutions. The full parallel configuration also enables the incorporation of multiple highly expressive models, such as Transformer-based models. 

\subsubsection{Selection of convolution layer}\label{sec:different-cnn}
The convolution layer is crucial for distilling information from the long historical sequence. In the main study, we utilized temporal convolutional network without padding to extract long-range dependency of historical information. We compare the performance of TCN, CNN and maxpooling methods within WEITS architecture by replacing the convolutional layer to corresponding configurations, with results shown in Table S5. 
\subsubsection{Computational cost comparison}
We present the comparison of computational costs between WEITS and the existing benchmarks. The experiments are conducted with the same input format as the data input in ETTm1 dataset. For existing models, we build them with the best hyper-parameter settings presented in their source codes. For WEITS-1, we use parameter settings illustrated in Section \ref{sec:train}. For WEITS-2, which incorporates Informer-based components, we keep all parameters and hyper-parameters as stated in the original paper. The results are calculated with ”ptflops” package \cite{ptflops}, with results shown in Table S7. WEITS can be configured to be computationally efficient while still yielding accurate results, and can also be used to fine-tune the output of large models, resulting in significant improvement with marginal computation cost.

\section{Discussion}
Experiment results have confirmed the effectiveness of signal enhancement for residual stacks and multi-resolution sampling mechanism for interpretable time series forecasting. WEITS is the first framework that infuses wavelet techniques into residual networks, which automatically produces interpretable stack outputs without manual configuration of each stack, and outperforms SoTA baselines in multiple tasks. Such innovative decomposition approach provides reliable insights to clients, boosting their understanding and confidence in high-stakes scenarios, such as financial prediction and disease modeling \citep{sun2021manifold}. 

There are also some interesting directions for future exploration. While WEITS exists as a uni-variate time series method, its potential can be extended to multivariate scenarios. Existing methods generalize into multivariate scenarios through simple tensor flattening (concatenation) \citep{nbeats}, which may underperform due to inability of efficiently capturing information from high -dimensional inputs, as the series after concatenation can be too long. This indicates a research potential of new mechanisms, such as stacked 1d temporal convolutional neural network, to effectively capture inter-series patterns. Moreover, Through stack-level integration of new Transformer-inspired architectures, the ability of long-horizon forecast may be further promoted. To summarize, WEITS provides a powerful, general and interpretable framework for time series forecast, and serves as a strong guidance to enlight future research in multi-variate, long-horizon scenarios.

\section*{Acknowledgments}
This work is supported by the National Key Research
and Development Program of China (Grant No.
2022YFB3103702).

% \bibliographystyle{apalike}
% \bibliography{sample}

\appendix % 切换到附录部分
\newpage
\onecolumn
\setcounter{section}{0}

\setcounter{table}{0}
\renewcommand{\thetable}{S\arabic{table}} % 设置表格计数器的显示格式为 S1, S2, S3...

\section{Data sources}\label{dataset}
We evaluate the performance of various long-term series forecasting algorithms using a diverse set of 4 datasets. Details of these datasets are listed below:
\begin{enumerate}
    \item Stock market data is an example of non-stationary data. At particular time there can be trends, cycles, random walks or combined effects \citep{patel2015predicting}. Therefore, high interpretability is crucial for understanding the changes of the market. We extracted the SP500 hourly data of 500 major companies \citep{uotila2009exploration} from 2015-01-01 to 2022-12-30. We aim to predict $H$ hours ahead utilizing a preceding historical window of 300 hours. In the dataset, starting points are randomly selected for each sample, while all samples do not have any overlapping time periods. We randomly split the datasets into train (70\%), validation (10\%) and test (20\%) splits following \cite{challu2022n}. Train and validation subsets are used for tuning and selecting optimal hyperparameters. Once the hyperparameters are determined, the model will be trained on the full train set, and the performance will be evaluated on the test subset. 
    \item The ETT dataset(Zhou et al., 2021) is a collection of load and oil temperature data from electricity transformers, captured at 15-minute intervals between July 2016 and July 2018. The dataset comprises four sub-datasets, namely ETTm1, ETTm2, ETTh1, and ETTh2, which correspond to two different transformers (labeled with 1 and 2) and two different resolutions (15 minutes and 1 hour). Each sub-dataset includes seven oil and load features of electricity transformers.
    \item The ILI dataset is a collection of weekly data on the ratio of patients exhibiting influenza-like symptoms to the total number of patients, as reported by the Centers for Disease Control and Prevention of the United States. The dataset spans the period from 2002 to 2021 and has been used in various studies related to influenza surveillance and analysis
\end{enumerate}

\section{Proof}

In this Appendix we prove the \emph{neural basis expansion approximation theorem} \cite{challu2022n} which can be applied to showcase the capability of our methodology to approximate for fixed horizons $t\in [0,1]$. We prove the case when the forecast function is a linear combination of basic functions, i.e., $g_{w,h}(\tau)=\theta_{w,h}\phi_{w,h}(\tau)=\theta_{w,h}\mathbbm{1}\{\tau \in [2^{-w}(h-1),2^{-w}h]\}$ are piece-wise constants and the inputs $y_{t-L:t}\in[0,1]$. The proof for more general functions is analogous.

\textbf{Lemma 1.} Let a function representing an infinite forecast horizon be $\mathcal{Y}: [0, 1] \rightarrow \mathbb{R}$ a square integrable function $\mathcal{L}^{2}([0,1])$. The forecast function $\mathcal{Y}$ can be arbitrarily well approximated by a linear combination of piecewise constants:
$$V_{w}=\{\phi_{w,h}(\tau) = \phi(2^{w}(\tau-h)) \;|\; w \in {\mathbb{Z}}, h\in2^{-w}\times[0,\dots,2^{w}]\} \quad $$
where $w \in \mathbb{N}$ controls the frequency/indicator's length and $h$ the time-location (knots) around which the indicator $\phi_{w,h}(\tau) = \mathbbm{1}\{\tau \in [2^{-w}(h-1),2^{-w}h]\}$ is active. That is, $\forall \epsilon >0$, there is a $w \in \mathbb{N}$ and $\hat{\mathcal{Y}}(\tau |y_{t-L:t})=\mathrm{Proj}_{V_{w}}(\mathcal{Y}(\tau|y_{t-L:t})) \in \mathrm{Span}(\phi_{w,h})$ such that

\begin{equation}
    \int_{[0,1]}|\mathcal{Y}(\tau) - \hat{\mathcal{Y}}(\tau)| d\tau =
    \int_{[0,1]} |\mathcal{Y}(\tau) - \sum_{w,h} \theta_{w,h} \phi_{w,h}(\tau) | d\tau \leq \epsilon %\quad \iff \quad \lim_{w\to\infty} |||| = 0
\end{equation}

\begin{proof}

The detailed proof of Lemma 1 can be traced back to Haar's classic work (1910). The indicator functions $V_{w} = \{\phi_{w,h}(\tau)\}$ are also referred in literature as Haar scaling functions or father wavelets in \cite{boggess2015first}. Denote the number of coefficients for the $\epsilon$-approximation $\hat{\mathcal{Y}}(\tau |y_{t-L:t})$ as $N_{\epsilon} = \sum^{w}_{i=0} 2^i=2^{w+1}-1$.

\end{proof}

Lemma 1 shows that a piecewise constant function can approximate a smooth and continuous function arbitrarily well when the discontinuity knots become dense. In Lemma 2, we show that the approximated function can be represented by a multi-resolution neural network with a finite number of  multi-resolution coefficients, which is a special case of the neural universal approximation theorem that states the approximation capacity of neural networks of arbitrary width \citep{hornik1991approximation}. 

\textbf{Lemma 2.}
Let a forecast mapping $\mathcal{Y}(\cdot\;|\;y_{t-L:t}): [0,1]^{L} \to \mathcal{L}^{2}([0,1])$ be $\epsilon$-approximated by $\hat{\mathcal{Y}}(\tau |y_{t-L:t})=\mathrm{Proj}_{V_{w}}(\mathcal{Y}(\tau|y_{t-L:t}))$, the projection to multi-resolution piecewise constants. If the relationship between $y_{t-L:t} \in [0,1]^{L}$ and $\theta_{w,h}$ varies smoothly, for instance $\theta_{w,h}:[0,1]^{L}\to \mathbb{R}$ is a K-Lipschitz function then for all $\epsilon > 0$ there exists a three-layer neural network $\hat{\theta}_{w,h}: [0, 1]^L \rightarrow \mathbb{R}$ with $O\left(L\right(\frac{K}{\varepsilon}\left)^L\right)$ neurons and $\mathrm{ReLU}$ activations such that 
\begin{equation}
\int_{[0, 1]^L} |\theta_{w,h}(y_{t-L:t}) - \hat{\theta}_{w,h}(y_{t-L:t})| dy_{t-L:t} \leq \epsilon
\end{equation}

% \vspace{5mm}
\textbf{Theorem 1.} Let a forecast mapping be

$\mathcal{Y}(\cdot \;|\; y_{t-L:t}): [0, 1]^L \rightarrow \mathcal{F}$, where the forecast functions $\mathcal{F}=\{\mathcal{Y}(\tau): [0,1] \to \mathbb{R}\}=\mathcal{L}^{2}([0,1])$ representing a continuous horizon, are square integrable. 

If the multi-resolution functions $V_{w}$ can arbitrarily approximate $\mathcal{L}^{2}([0,1])$, and the projection $\mathrm{Proj}_{V_{w}}(\mathcal{Y}(\tau))$ varies smoothly on $y_{t-L:t}$, then the forecast mapping $\mathcal{Y}(\cdot \;|\; y_{t-L:t})$ can be arbitrarily approximated by a neural network learning a finite number of  multi-resolution coefficients $\hat{\theta}_{w,h}$. Denote the proportion of wavelet infusion as $\alpha$, and the corresponding forecast as $\tilde{\mathcal{Y}}_\alpha$. As a special case, when $\alpha\rightarrow 0$, the model converges to existing methodology of \cite{challu2022n}.

$\forall \epsilon>0, \alpha \in [0,1]$, 
\begin{align}
% \nonumber
\begin{split}
    &\min\limits_{\alpha \in[0,1]}\int |\mathcal{Y}(\tau \;|\; y_{t-L:t}) -  \tilde{\mathcal{Y}}_\alpha(\tau \;|\; y_{t-L:t}) | d\tau \\
    \leq &\int |\mathcal{Y}(\tau \;|\; y_{t-L:t}) -  \tilde{\mathcal{Y}}_{\alpha=0}(\tau \;|\; y_{t-L:t}) | d\tau \qquad\qquad\qquad\qquad\qquad \\
    =& \int |\mathcal{Y}(\tau \;|\; y_{t-L:t}) 
    -\sum_{w,h} \hat{\theta}_{w,h}(y_{t-L:t}) \phi_{w,h}(\tau)| d\tau \leq \epsilon
\end{split}
\end{align}

% \begin{equation}
%     \int |\mathcal{Y}(\tau \;|\; y_{t-L:t}) - \tilde{\mathcal{Y}}(\tau \;|\; y_{t-L:t}) | d\tau = \int |\mathcal{Y}(\tau \;|\; y_{t-L:t}) -\sum_{w,h} \hat{\theta}_{w,h}(y_{t-L:t}) \phi_{w,h}(\tau)| d\tau \leq \epsilon
% \end{equation}

% \clearpage
\begin{proof} For simplicity of the proof, we will omit the conditional lags $y_{t-L:t}$. Using both the neural approximation $\tilde{\mathcal{Y}}$ from Lemma 2, and Haar's approximation $\hat{\mathcal{Y}}$ from Lemma 1, 
\begin{align*}
    \int |\mathcal{Y}(\tau) -\tilde{\mathcal{Y}}_{\alpha}(\tau)| d\tau 
    &= 
    \int |(\mathcal{Y}(\tau) - \hat{\mathcal{Y}}(\tau))+(\hat{\mathcal{Y}}(\tau)-\tilde{\mathcal{Y}}_{\alpha}(\tau))| d\tau 
\end{align*}

By the triangular inequality:
\begin{align*}
\begin{split}
    \int |\mathcal{Y}(\tau) -\tilde{\mathcal{Y}}_{\alpha}(\tau)| d\tau & \leq \int |\mathcal{Y}(\tau) - \hat{\mathcal{Y}}(\tau)| \\
    & +
    |\sum_{w,h} \theta_{w,h} \phi_{w,h}(\tau) - \sum_{w,h} \hat{\theta}_{w,h} \phi_{w,h}(\tau)| d \tau %\\
\end{split}
\end{align*}

By a special case of Fubini's theorem
\begin{align*}
    \int |\mathcal{Y}(\tau) -\tilde{\mathcal{Y}}_{\alpha}(\tau)| d\tau \leq 
    \qquad\qquad\qquad\qquad\qquad\qquad\qquad\quad \\
    \int 
    |\mathcal{Y}(\tau) - \sum_{w,h} \hat{\mathcal{Y}}(\tau)| d \tau 
    +
    \sum_{w,h} \int_{\tau} 
    |(\theta_{w,h} - \hat{\theta}_{w,h}) \phi_{w,h}(\tau)| d \tau %\\
\end{align*}    

Using positivity and bounds of the indicator functions
\begin{align*}
    \int |\mathcal{Y}(\tau) -\tilde{\mathcal{Y}}_{\alpha}(\tau)| d\tau 
    \leq \qquad\qquad\qquad\qquad\qquad\qquad\qquad\quad\\
    \int_{\tau} 
    |\mathcal{Y}(\tau) - \sum_{w,h} \hat{\mathcal{Y}}(\tau)| d \tau 
    +
    \sum_{w,h} |\theta_{w,h} - \hat{\theta}_{w,h}| \int_{\tau} \phi_{w,h}(\tau) d \tau \\
    < \int_{\tau} 
    |\mathcal{Y}(\tau) - \sum_{w,h} \hat{\mathcal{Y}}(\tau)| d \tau
    +
    \sum_{w,h} |\theta_{w,h} - \hat{\theta}_{w,h}| 
\end{align*} 

To conclude we use the both arbitrary approximations from the Haar projection and the approximation to the finite multi-resolution coefficients
\begin{align*}
    \int |\mathcal{Y}(\tau) -\tilde{\mathcal{Y}}_{\alpha}(\tau)| d\tau    
    \leq \qquad\qquad\qquad\qquad\qquad\qquad\qquad\quad\\
    \int |\mathcal{Y}(\tau) - \hat{\mathcal{Y}}(\tau)| d \tau 
    +
    \sum_{w,h} |\theta_{w,h} - \hat{\theta}_{w,h}| %\\
    \leq \epsilon_{1} + N_{\epsilon_{1}} \epsilon_{2} \; \leq \epsilon
\end{align*}
\end{proof}

\section{Tables}
\begin{table*}[h!]
\footnotesize
\label{table1}
\caption{Supplement to the comparison results in Table 1.}
\begin{center}
\resizebox{\textwidth}{!}{\begin{tabular}{cc|cccccccccccccc}
\toprule
 \multicolumn{2}{c}{Model}&\multicolumn{2}{c}{WEITS-3}&\multicolumn{2}{c}{WEITS-4}&\multicolumn{2}{c}{PatchTST}&\multicolumn{2}{c}{SCINet}&\multicolumn{2}{c}{DLinear}&\multicolumn{2}{c}{FEDformer}&\multicolumn{2}{c}{Transformer} \\
 Dataset &  H &MSE&MAE&MSE&MAE&MSE&MAE&MSE&MAE&MSE&MAE&MSE&MAE&MSE&MAE \\
 \hline 
\multirow{4}{*}{\rotatebox{90}{SP-500}}
&6  &0.208&0.275&0.222&0.301&0.211&0.267&0.227&0.281&0.229&0.290&0.231&0.326&0.289&0.351\\
&12 &0.249&0.316&0.240&0.336&0.236&0.320&0.240&0.319&0.243&0.332&0.281&0.358&0.303&0.378\\
&18 &0.278&0.340&0.295&0.362&0.254&0.327&0.261&0.341&0.267&0.358&0.304&0.395&0.348&0.441\\
&24 &0.330&0.379&0.345&0.390&0.329&0.391&0.338&0.372&0.369&4.237&0.468&0.492&0.607&0.583\\
\midrule
\multirow{3}{*}{\rotatebox{90}{ETTh1}}
&24 &0.263&0.321&0.287&0.334&0.258&0.290&0.269&0.287&0.281&0.306&0.301&0.349&0.393&0.471\\ 
&36 &0.316&0.407&0.320&0.389&0.301&0.351&0.328&0.369&0.347&0.399&0.380&0.402&0.476&0.470\\ 
&48 &0.439&0.490&0.448&0.526&0.399&0.460&0.408&0.471&0.427&0.501&0.475&0.559&0.528&0.517\\ 
\midrule
\multirow{3}{*}{\rotatebox{90}{ETTh2}}
&24  &0.208&0.244&0.241&0.289&0.203&0.257&0.206&0.261&0.226&0.293&0.259&0.331&0.309&0.392\\
&36  &0.257&0.301&0.327&0.371&0.224&0.296&0.219&0.301&0.251&0.302&0.271&0.349&0.378&0.466\\
&48  &0.329&0.372&0.395&0.419&0.325&0.410&0.337&0.424&0.348&0.429&0.398&0.440&0.528&0.515\\
\midrule
\multirow{3}{*}{\rotatebox{90}{ILI}}
&24 &2.160&0.999&2.237&1.154&1.850&0.897&1.825&0.912&2.085&0.932&2.109&0.992&2.294&1.087\\
&36 &2.285&1.013&2.394&1.109&2.112&0.939&2.166&0.960&2.124&0.955&2.188&1.031&2.530&1.280      \\
&48 &2.360&1.117&2.405&1.240&2.256&0.962&2.249&1.001&2.307&1.078&2.390&1.109&3.098&1.684      \\
\bottomrule
\end{tabular}}
\end{center}
\end{table*}

\begin{table*}
\footnotesize
\label{table1}
\caption{The comparison of accuracy at
different infusion levels. From the table, it is evident that the effect is most pronounced when 
$\alpha$ is set between 0.3 and 0.45. This setting is also consistent with our model in the comparison results.}
\begin{center}
\resizebox{\textwidth}{!}{ 
\begin{tabular}{cc|ccccccccccccccccccc}
\toprule
 \multicolumn{2}{c}{$\alpha$}&\multicolumn{2}{c}{0}&\multicolumn{2}{c}{0.15}&\multicolumn{2}{c}{0.3}&\multicolumn{2}{c}{0.45}&\multicolumn{2}{c}{0.6}&\multicolumn{2}{c}{0.75}&\multicolumn{2}{c}{0.9}&\multicolumn{2}{c}{1} \\
 Dataset & H &MSE&MAE&MSE&MAE&MSE&MAE&MSE&MAE&MSE&MAE&MSE&MAE&MSE&MAE&MSE&MAE \\
 \hline
\multirow{4}{*}{\rotatebox{90}{SP-500}}
&6  
&0.241&0.337&0.227&0.283&0.194&0.276&0.203&0.265&0.235&0.289&0.241&0.279&0.302&0.341&0.324&0.381\\
&12 
&0.305&0.367&0.242&0.311&0.229&0.288&0.231&0.293&0.249&0.315&0.348&0.407&0.389&0.453&0.427&0.509\\
&18 
&0.426&0.497&0.274&0.335&0.261&0.322&0.263&0.309&0.280&0.333&0.378&0.425&0.459&0.462&0.498&0.527\\
&24 
&0.482&0.541&0.345&0.387&0.324&0.368&0.332&0.381&0.347&0.416&0.428&0.469&0.501&0.517&0.529&0.542\\
\midrule
\multirow{3}{*}{\rotatebox{90}{ETTh1}}
&24
&0.291&0.340&0.269&0.321&0.245&0.291&0.241&0.307&0.272&0.329&0.299&0.331&0.316&0.357&0.340&0.398\\ 
&36
&0.402&0.433&0.337&0.391&0.302&0.360&0.294&0.356&0.325&0.387&0.346&0.360&0.398&0.421&0.427&0.486\\
&48
&0.528&0.595&0.478&0.505&0.428&0.477&0.451&0.492&0.472&0.506&0.526&0.551&0.603&0.597&0.630&0.658\\
\midrule
\multirow{3}{*}{\rotatebox{90}{ETTh2}}
&24  
&0.290&0.336&0.247&0.292&0.199&0.241&0.210&0.244&0.249&0.278&0.315&0.341&0.349&0.381&0.368&0.417\\
&36  
&0.397&0.468&0.269&0.319&0.240&0.287&0.232&0.270&0.306&0.327&0.355&0.390&0.419&0.498&0.461&0.526\\
&48 
&0.487&0.561&0.356&0.412&0.321&0.349&0.339&0.380&0.379&0.405&0.417&0.429&0.520&0.597&0.582&0.640\\
\midrule
\multirow{3}{*}{\rotatebox{90}{ILI}}
&24 
&2.275&1.322&2.214&1.192&1.978&0.993&1.904&0.982&2.101&1.062&2.202&1.164&2.376&1.408&2.416&1.434\\
&36 
&2.261&1.529&2.568&1.309&2.271&1.037&2.004&1.026&2.354&1.098&2.689&1.280&2.710&1.551&2.801&1.629\\
&48 
&3.094&1.685&2.817&1.426&2.303&1.089&2.208&1.175&2.577&1.278&2.841&1.459&3.075&1.604&3.271&1.689\\
\bottomrule
\end{tabular}}
\end{center}
\end{table*}

\begin{table*}
\footnotesize
\label{table1}
\caption{Injecting random multiplicative noise for robustness testing. From the table, it can be seen that WEITS has good robustness and is relatively insensitive to noise compared to other models. We obtained the best results in most experimental settings when setting the stack to 6, and increasing the number of stacks did not result in better accuracy with a larger model size.}
\begin{center}
\
\begin{tabular}{cc|cccccc}
\toprule
 \multicolumn{2}{c}{The degree of noise}&\multicolumn{2}{c}{2.5$\%$}&\multicolumn{2}{c}{5$\%$}&\multicolumn{2}{c}{7.5$\%$} \\
 Dataset & Model &MSE&MAE&MSE&MAE&MSE&MAE \\
 \hline
\multirow{3}{*}{\rotatebox{90}{SP-500}}
&WEITS-1 &\textbf{0.257}&\textbf{0.319}&\textbf{0.318}&\textbf{0.345}&0.387&\textbf{0.426}\\
&N-HiTS &0.259&0.326&0.336&0.350&\textbf{0.382}&0.477\\
&DLinear &0.266&0.357&0.375&0.391&0.429&0.490\\
\midrule
\multirow{3}{*}{\rotatebox{90}{ETTh1}}
&WEITS-1 &\textbf{0.271}&\textbf{0.309}&\textbf{0.299}&\textbf{0.331}&\textbf{0.348}&\textbf{0.397}\\
&N-HiTS &0.330&0.328&0.367&0.352&0.390&0.409\\
&DLinear &0.321&0.361&0.321&0.345&0.401&0.438\\ 
\midrule
\multirow{3}{*}{\rotatebox{90}{ETTh2}}
&WEITS-1 &\textbf{0.223}&\textbf{0.242}&\textbf{0.246}&\textbf{0.285}&\textbf{0.236}&\textbf{0.287}\\
&N-HiTS &0.229&0.280&0.259&0.291&0.288&0.318\\
&DLinear &0.244&0.306&0.271&0.327&0.359&0.380\\ 
\midrule
\multirow{3}{*}{\rotatebox{90}{ILI}}
&WEITS-1 &\textbf{2.204}&\textbf{0.983}&\textbf{2.478}&\textbf{1.209}&\textbf{2.691}&\textbf{1.474}\\
&N-HiTS &2.307&1.025&2.582&1.287&2.732&1.503\\
&DLinear &2.341&1.197&2.549&1.301&2.870&1.529\\ 
\bottomrule
\end{tabular}
\end{center}
\end{table*}

\begin{table*}
\footnotesize
\label{table1}
\caption{The relation between the number of wavelet stacks and accuracy. We obtained the best results in most experimental settings when setting the stack to 6, and increasing the number of stacks did not result in better accuracy with a larger model size.}
\begin{center}
\begin{tabular}{cc|cccccccc}
\toprule
\multicolumn{2}{c}{N of stacks}&\multicolumn{2}{c}{3}&\multicolumn{2}{c}{4}&\multicolumn{2}{c}{5}&\multicolumn{2}{c}{6} \\
Dataset & H &MSE&MAE&MSE&MAE&MSE&MAE&MSE&MAE \\
\hline
\multirow{4}{*}{\rotatebox{90}{SP-500}}
&6  &0.493&0.527&0.317&0.387&0.190&0.278&\textbf{0.183}&\textbf{0.261} \\
&12 &0.523&0.596&0.340&0.461&0.248&0.311&\textbf{0.225}&\textbf{0.282} \\
&18 &0.749&0.692&0.408&0.487&0.274&0.312&\textbf{0.249}&\textbf{0.305} \\
&24 &0.889&0.827&0.498&0.509&0.341&0.398&\textbf{0.319}&\textbf{0.354} \\
\midrule
\multirow{3}{*}{\rotatebox{90}{ETTh1}}
&24 &0.513&0.489&0.359&0.367&0.263&0.289& \textbf{0.244} & \textbf{0.287} \\ 
&36 &0.754&0.669&0.412&0.441&0.305&\textbf{0.340}& \textbf{0.296} & 0.347 \\ 
&48 &1.084&0.926&0.562&0.659&0.421&0.467& \textbf{0.401} & \textbf{0.442} \\ 
\midrule
\multirow{3}{*}{\rotatebox{90}{ETTh2}}
&24 &0.348&0.373&0.294&0.329&0.208&0.249&\textbf{0.192}&\textbf{0.234} \\
&36 &0.402&0.469&0.325&0.409&0.269&0.308&\textbf{0.228}&\textbf{0.260} \\
&48 &0.528&0.632&0.461&0.457&0.342&0.388&\textbf{0.319}&\textbf{0.359} \\
\midrule
\multirow{3}{*}{\rotatebox{90}{ILI}}
&24 &2.297&1.446&2.203&1.159&1.998&1.035&\textbf{1.814}&\textbf{0.903} \\
&36 &2.563&1.509&2.414&1.398&2.130&1.238&\textbf{2.071}&\textbf{0.937} \\
&48 &2.970&1.782&2.689&1.697&2.281&1.460&\textbf{2.134}&\textbf{0.894} \\
\bottomrule
\end{tabular}
\end{center}
\end{table*}

\begin{table*}
\footnotesize
\label{table1}
\caption{The relationship between ensemble size and accuracy. The use of the ensemble model leads to a noticeable improvement in accuracy. In our comparison results. We scaled the ensemble model up to 40, which shows a significant performance boost compared to smaller-scale ensembles.}
\begin{center}
\
\begin{tabular}{cc|ccccccccccccccc}
\toprule
 \multicolumn{2}{c}{ensemble size}&\multicolumn{2}{c}{10}&\multicolumn{2}{c}{20}&\multicolumn{2}{c}{40} \\
 Dataset & H &MSE&MAE&MSE&MAE&MSE&MAE \\
 \hline
\multirow{4}{*}{\rotatebox{90}{SP-500}}
&6  &0.254&0.326&0.207&0.297&\textbf{0.183}&\textbf{0.261}\\
&12 &0.297&0.363&0.261&0.313&\textbf{0.225}&\textbf{0.282}\\
&18 &0.345&0.417&0.272&0.338&\textbf{0.249}&\textbf{0.305}\\
&24 &0.501&0.527&0.368&0.463&\textbf{0.319}&\textbf{0.354}\\
\midrule
\multirow{3}{*}{\rotatebox{90}{ETTh1}}
&24 &0.319&0.398&0.294&0.356& \textbf{0.244} & \textbf{0.287}\\ 
&36 &0.405&0.441&0.346&0.391& \textbf{0.296} & \textbf{0.347}\\ 
&48 &0.764&0.608&0.528&0.613& \textbf{0.401} & \textbf{0.442}\\ 
\midrule
\multirow{3}{*}{\rotatebox{90}{ETTh2}}
&24  &0.304&0.328&0.241&0.275&\textbf{0.192}&\textbf{0.234}\\
&36  &0.358&0.420&0.309&0.344&\textbf{0.228}&\textbf{0.260}\\
&48  &0.506&0.488&0.427&0.451&\textbf{0.319}&\textbf{0.359}\\
\midrule
\multirow{3}{*}{\rotatebox{90}{ILI}}
&24 &2.987&1.556&2.560&1.294&\textbf{1.814}&\textbf{0.903}\\
&36 &3.218&1.640&2.743&1.327&\textbf{2.071}&\textbf{0.937}\\
&48 &3.856&1.679&2.810&1.378&\textbf{2.134}&\textbf{0.894}\\
\bottomrule
\end{tabular}
\end{center}
\end{table*}

\begin{table*}
\footnotesize
\label{table1}
\caption{Comparison the performance of different convolutional layer within WEITS architecture. After replacing different convolutional layers, it can be observed that the use of DCN results in a significant performance improvement compared to other types of convolutional layers.}
\begin{center}
\
\begin{tabular}{cc|ccccccccccccccc}
\toprule
 \multicolumn{2}{c}{layer}&\multicolumn{2}{c}{DCN}&\multicolumn{2}{c}{CNN}&\multicolumn{2}{c}{MaxPooling}&\multicolumn{2}{c}{AveragePooling} \\
 Dataset & H &MSE&MAE&MSE&MAE&MSE&MAE&MSE&MAE \\
 \hline
\multirow{4}{*}{\rotatebox{90}{SP-500}}
&6  &\textbf{0.183}&\textbf{0.261}&0.214&0.291&0.367&0.408&0.321&0.379\\
&12 &\textbf{0.225}&\textbf{0.282}&0.276&0.301&0.391&0.416&0.389&0.427\\
&18 &\textbf{0.249}&\textbf{0.305}&0.314&0.345&0.456&0.502&0.653&0.742\\
&24 &\textbf{0.319}&\textbf{0.354}&0.411&0.408&0.881&0.807&0.812&0.796\\
\midrule
\multirow{3}{*}{\rotatebox{90}{ETTh1}}
&24 & \textbf{0.244} & \textbf{0.287} &0.340&0.407&0.426&0.508&0.405&0.467\\ 
&36 & \textbf{0.296} & \textbf{0.347} &0.387&0.458&0.489&0.521&0.452&0.522\\ 
&48 & \textbf{0.401} & \textbf{0.442} &0.509&0.539&0.591&0.587&0.583&0.691\\ 
\midrule
\multirow{3}{*}{\rotatebox{90}{ETTh2}}
&24  &\textbf{0.192}&\textbf{0.234}&0.221&0.296&0.241&0.308&0.234&0.317\\
&36  &\textbf{0.228}&\textbf{0.260}&0.328&0.361&0.289&0.362&0.294&0.356\\
&48 &\textbf{0.319}&\textbf{0.359}&0.459&0.427&0.501&0.481&0.479&0.523\\
\midrule
\multirow{3}{*}{\rotatebox{90}{ILI}}
&24 &\textbf{1.814}&\textbf{0.903}&2.008&1.003&2.856&1.228&2.767&1.394\\
&36 &\textbf{2.071}&\textbf{0.937}&2.206&1.098&3.054&1.304&2.980&1.470\\
&48 &\textbf{2.134}&\textbf{0.894}&2.287&1.153&3.378&1.344&3.296&1.556\\
\bottomrule
\end{tabular}
\end{center}
\end{table*}

\begin{table*}
\footnotesize
\label{table1}
\caption{The comparison of computational costs between WEITS and the existing benchmarks.}
\begin{center}
\
\resizebox{\textwidth}{!}{\begin{tabular}
{c|cccccccccc}
\toprule
 Model&WEITS-1&WEITS-1&Autoformer&Autoformer&PatchTST&PatchTST&Informer&Informer 
 
 \\
 \hline
 &(FLOPS)&(Params)&(FLOPS)&(Params)&(FLOPS)&(Params)&(FLOPS)&(Params)
 
 \\
 \midrule
H=160&7.26G&10.6M&11.2G&15.5M&5.28G&5.34M&9.72G&11.3M
\\
H=320&9.24G&10.6M&12.5G&15.5M&5.37G&9.88M&11.0G&11.3M
\\
H=480&11.87G&10.6M&13.9G&15.5M&5.46G&13.9M&11.8G&11.3M
\\
\bottomrule
\end{tabular}}
\end{center}
\end{table*}

\begin{table*}
\footnotesize
\label{table1}
\caption{The performance of WEITS on multivariate time series forecasting.}
\begin{center}
\
\resizebox{\textwidth}{!}{\begin{tabular}{cc|ccccccccccccccc}
\toprule
 \multicolumn{2}{c}{Model}&\multicolumn{2}{c}{WEITS-1}&\multicolumn{2}{c}{WEITS-2}&\multicolumn{2}{c}{N-HiTs}&\multicolumn{2}{c}{N-Beats}&\multicolumn{2}{c}{DeepAR}&\multicolumn{2}{c}{Informer} \\
 Dataset & H &MSE&MAE&MSE&MAE&MSE&MAE&MSE&MAE&MSE&MAE&MSE&MAE \\
 \hline
\multirow{3}{*}{\rotatebox{90}{ETTm2}}
&48 
&\textbf{0.170}&\textbf{0.216}&0.208&0.245&0.183&0.221&0.287&0.320&0.337&0.370&0.342&0.378\\
&96
&2.074&0.249&0.225&0.254&\textbf{0.198}&\textbf{0.219}&0.351&0.397&0.493&0.509&0.502&0.527\\
&144
&0.258&\textbf{0.308}&0.276&0.324&\textbf{0.240}&0.312&0.532&0.584&0.649&0.671&0.608&0.644\\
\midrule
\multirow{3}{*}{\rotatebox{90}{ECL}}
&48 
&0.204&0.241&0.236&0.298&\textbf{0.197}&\textbf{0.249}&0.311&0.385&0.327&0.391&0.342&0.390\\
&96 
&\textbf{0.221}&0.283&0.258&0.319&0.231&\textbf{0.271}&0.356&0.402&0.412&0.521&0.397&0.504\\
&144
&0.294&0.328&0.316&0.347&\textbf{0.244}&\textbf{0.293}&0.436&0.507&0.489&0.551&0.431&0.612 \\
\midrule
\multirow{3}{*}{\rotatebox{90}{Weather}}
&48
&\textbf{0.151}&\textbf{0.228}&0.167&0.249&0.158&0.239&0.281&0.336&0.328&0.399&0.378&0.465 \\
&96
&0.179&0.245&0.194&0.261&\textbf{0.160}&\textbf{0.207}&0.343&0.422&0.408&0.519&0.431&0.568 \\
&144
&0.244&0.317&0.242&0.308&\textbf{0.211}&\textbf{0.268}&0.389&0.461&0.508&0.602&0.445&0.627\\
\bottomrule
\end{tabular}}
\end{center}
\end{table*}

\vfill

\end{document}